\documentclass[preprint,12pt]{elsarticle}

\usepackage{amssymb}
\usepackage{amsmath}

\usepackage{booktabs}
\usepackage[utf8]{inputenc}
\usepackage{amsmath, amssymb}
\usepackage{hyperref}
\usepackage{pifont}
\usepackage{xcolor}

\DeclareMathOperator{\mape}{\textsc{mape}}
\DeclareMathOperator{\mpe}{\textsc{mpe}}
\DeclareMathOperator{\ape}{\textsc{ape}}
\DeclareMathOperator{\medape}{\textsc{medape}}

\DeclareMathOperator{\rmse}{\textsc{rmse}}

\DeclareMathOperator{\iqr}{\textsc{iqr}}

\journal{Applied Soft Computing}

\begin{document}

\begin{frontmatter}



\title{Enhanced N-BEATS for Mid-Term Electricity Demand Forecasting}


\author[a]{Mateusz Kasprzyk} 
\ead{aeris451@outlook.jp}
\author[b]{Paweł Pełka}
\ead{pawel.pelka@pcz.pl}
\author[c]{Boris N. Oreshkin\corref{cor2}}
\ead{boris.oreshkin@gmail.com}
\author[b]{Grzegorz Dudek\corref{cor1}}
\ead{grzegorz.dudek@pcz.pl}
\cortext[cor1]{Corresponding author.}
\cortext[cor2]{This work does not relate to author's position at Amazon.}

\affiliation[a]{organization={Faculty of Computer Science and Artificial Intelligence, Czestochowa University of Technology},
            addressline={ul. Dabrowskiego 73}, 
            city={Czestochowa},
            postcode={42-201}, 
            country={Poland}}

\affiliation[b]{organization={Faculty of Electrical Engineering,
Czestochowa University of Technology},
            addressline={Al. Armii Krajowej 17}, 
            city={Czestochowa},
            postcode={42-201}, 
            country={Poland}}

\affiliation[c]{organization={Amazon Science},
            addressline={120 Bremner Blvd}, 
            city={Toronto},
            postcode={M5J 0A8}, 
            state={ON},
            country={Canada}}            

\begin{abstract}
This paper presents an enhanced N-BEATS model, N-BEATS*, for improved mid-term electricity load forecasting (MTLF). Building on the strengths of the original N-BEATS architecture, which excels in handling complex time series data without requiring preprocessing or domain-specific knowledge, N-BEATS* introduces two key modifications. (1) A novel loss function -- combining pinball loss based on MAPE with normalized MSE, the new loss function allows for a more balanced approach by capturing both $L_1$ and $L_2$ loss terms. (2) A modified block architecture -- the internal structure of the N-BEATS blocks is adjusted by introducing a destandardization component to harmonize the processing of different time series, leading to more efficient and less complex forecasting tasks.
Evaluated on real-world monthly electricity consumption data from 35 European countries, N-BEATS* demonstrates superior performance compared to its predecessor and other established forecasting methods, including statistical, machine learning, and hybrid models. N-BEATS* achieves the lowest MAPE and RMSE, while also exhibiting the lowest dispersion in forecast errors.
\end{abstract}


\end{frontmatter}




\section{Introduction}

Mid-term load forecasting (MTLF) is essential for power system operators and planners, as it involves predicting electricity demand over a time horizon of several weeks to a year. Accurate MTLF supports informed decision-making across multiple aspects of power system management, including power plant scheduling, infrastructure expansion, market operations, and maintaining grid reliability and security. By anticipating future demand, utilities can optimize maintenance schedules, secure fuel supplies, and plan necessary capacity additions. Additionally, accurate forecasts across various time horizons are crucial for ensuring grid stability by maintaining the balance between supply and demand. Furthermore, precise forecasting enables strategic decision-making in energy markets, guiding the timing of electricity purchases and sales. In summary, load forecasting serves as a cornerstone for efficient, reliable, and resilient power system operations.

\subsection{Related Work}

MTLF methodologies have evolved significantly over the years, encompassing a wide range of techniques from traditional statistical approaches to cutting-edge artificial intelligence solutions. Recent research in this area has focused on improving accuracy and robustness, particularly in the face of increasing uncertainty and complexity in energy systems.

MTLF strategies generally fall into two categories: conditional and autonomous modeling~\cite{Ghi06}. Conditional modeling integrates broader economic and infrastructural contexts, utilizing variables such as economic indicators and power grid characteristics. Autonomous modeling, on the other hand, relies primarily on historical consumption data, temperature patterns, and seasonality factors, making it more suitable for economies with stable energy demand patterns~\cite{Pei11}.

Historically, classical statistical methods like ARIMA and exponential smoothing (ETS) dominated the MTLF landscape~\cite{Bas23,Gup20}. However, their limitations in capturing non-linear relationships and adapting to complex patterns prompted researchers to explore more sophisticated approaches~\cite{Sug11}.
Machine learning offered enhanced adaptability and the ability to capture intricate patterns. Examples include support vector machines \cite{Li22}, neural networks (NNs)~\cite{Sha24}, and fuzzy systems~\cite{Ahm19}, which significantly improved forecast accuracy in scenarios where traditional methods fell short.

MTLF time series exhibit significant seasonal patterns, prompting researchers to employ pattern-based methods and initial preprocessing before applying forecasting models. This approach has proven effective in both classical \cite{Pel23} and neural network \cite{Dud20} methodologies.

The advent of deep learning (DL) marked a paradigm shift in MTLF. DL architectures, with their ability to leverage massive datasets and extract complex patterns, overcame many limitations of classical NNs. Recurrent NNs such as Long Short-Term Memory (LSTM) and Gated Recurrent Unit (GRU) networks demonstrated remarkable efficacy in handling long-term dependencies in time series data~\cite{Yan18,Li23}. They are combined with classical methods to preprocess complex data.    
For instance, Bedi and Toshniwal~\cite{Bed19} proposed a DL framework that integrates Empirical Mode Decomposition (EMD) with LSTM for electric load forecasting, showing promising results in handling non-stationary and non-linear load data. Similarly, Dudek et al.~\cite{Dud21} introduced a hybrid hierarchical model for MTLF, which integrates Exponential Smoothing (ETS) with advanced LSTM networks and ensemble techniques, showcasing improved forecasting accuracy.

Recent years have seen the emergence of even more advanced DL architectures. 
The integration of attention mechanisms into DL models has advanced forecasting capabilities by enabling models to prioritize the most relevant parts of the input sequence. Attention-based architectures, such as the Transformer model, excel in capturing long-range dependencies within data, albeit with increased computational demands~\cite{Vas17,Li19}. Notably, in~\cite{Li23a}, a transformer-based MTLF model is proposed, capable of generating probabilistic forecasts with improved interpretability and the ability to handle data at low temporal resolutions.

A notable recent development is the N-BEATS architecture, which has achieved state-of-the-art performance across various forecasting tasks~\cite{Ore19}. Initially designed for general time series applications, N-BEATS has been successfully tailored for MTLF, showcasing its adaptability and effectiveness. In a study by Oreshkin et al.~\cite{Ore21}, the N-BEATS model demonstrated superior performance in MTLF, outperforming both traditional and contemporary forecasting methods while maintaining computational efficiency. This study further refined N-BEATS to align with the specific requirements of MTLF, leading to increased forecast accuracy and robustness.

Another emerging trend is the focus on interpretability in MTLF models. Baur et al.~\cite{Bau24} reviews literature on explainable and interpretable machine learning methods for electric load forecasting, identifying trends and techniques to improve forecast transparency and interpretability. 
 An example of an MTLF model with interpretability features is presented in \cite{Li23a}, where the transformer model is capable of explaining the contribution of each input feature to the predicted load at different times of the day. Such capabilities not only improve trust in the model's predictions but also provide valuable insights for decision-making in energy management.

In conclusion, the field of MTLF has undergone a significant transformation, progressing from classical statistical methods through early machine learning applications to advanced deep learning architectures and hybrid models. Recent studies, such as those on N-BEATS in MTLF, interpretable DL models, and ensemble forecasting methods~\cite{Cha23}, demonstrate that this field continues to evolve, offering increasingly accurate and flexible tools for load forecasting while addressing crucial aspects like interpretability and computational efficiency.

\subsection{Motivation and Contributions}

N-BEATS is one of the most advanced machine learning models for time series forecasting, incorporating unique features that enhance its effectiveness and flexibility \cite{Ore19}. It utilizes a modular block structure, where each block is specifically designed to capture distinct components of the data, enabling the model to decompose forecasts into patterns. Additionally, N-BEATS includes backward and forward residual connections between blocks, allowing it to learn complex temporal patterns through iterative refinement, which improves both predictive accuracy and convergence speed.

The architecture of N-BEATS is designed to handle complex time series data without requiring specialized preprocessing or feature engineering, making it highly adaptable across various forecasting tasks. 
As a purely data-driven model, N-BEATS does not rely on domain-specific assumptions or prior knowledge, which enables it to adapt seamlessly to diverse datasets without substantial modifications, supporting robust predictions for both short- and long-term forecasting horizons.

Most existing models for MTLF primarily focus on enhancing forecasting accuracy, while neglecting the potential bias in the forecasts, which plays a crucial role in MTLF. The N-BEATS implementation in \cite{Ore21} addresses this gap by introducing an effective mechanism to control forecasting bias using the pinball-MAPE loss function, demonstrating its effectiveness on real-world data.

Given these strengths, we selected N-BEATS to address our MTLF problem, introducing modifications to further enhance its accuracy. Our first enhancement involves designing a tailored loss function, building on prior research \cite{Ore21} that demonstrated the substantial benefits of optimized loss functions in forecasting accuracy. We refine this approach here to better align with MTLF requirements. The second modification involves adjusting the internal architecture of the blocks to better suit different forecasting tasks, further improving the model’s performance. 

Our contributions can be summarized as follows:

\begin{enumerate}
    \item {Novel Loss Function for MTLF}: We introduce a new loss function that combines pinball loss based on MAPE with normalized MSE. This formulation captures both $L_1$ and $L_2$ loss terms, allowing for a more balanced approach. Additionally, the pinball loss aspect enables control over forecast bias and can be used for forecasting quantiles, further enhancing the model’s flexibility.
   
    \item {Novel Block Architecture for N-BEATS}:  
    We propose a modification to the internal architecture of the N-BEATS blocks by introducing a destandardization component. This component harmonizes the different time series processed by the blocks, making the forecasting tasks they solve less complex and more efficient.
    
    \item{Empirical Results for MTLF}:
     We empirically demonstrate, using real-world data from 35 European countries, that the proposed N-BEATS* model significantly outperforms its predecessor as well as well-established statistical and state-of-the-art machine learning methods in terms of forecasting accuracy.
\end{enumerate}

The rest of the paper is organized as follows. Section~\ref{DFP} provides an overview of the data and formulates the forecasting problem. Section~\ref{FM} introduces the proposed N-BEATS* model for MTLF. The experimental framework used to evaluate the model's performance is detailed in Section~\ref{EX}. Section~\ref{DS} explores the model's innovations, practical implications, and limitations, while also outlining potential directions for future research. Finally, Section~\ref{CN} presents the conclusions of the study.

\section{Data and Forecasting Problem} \label{DFP}

\begin{figure}[t]
\centering
\includegraphics[width=\textwidth]{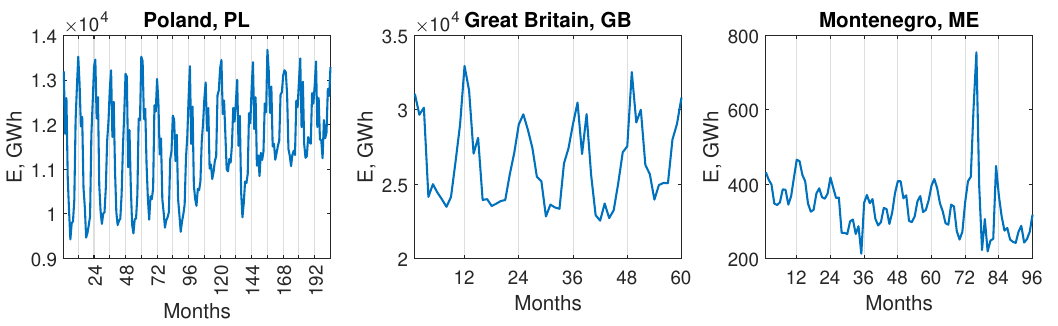}
\caption{Examples of monthly electricity demand time series.} \label{fig0}
\end{figure}

Monthly electricity demand time series exhibit intricate dynamics characterized by a non-linear trend, seasonal patterns, and stochastic variations. Examples of these time series are shown in Fig. \ref{fig0}, with more detailed analyses and visualizations available in \cite{Ore21}, \cite{Dud21} and \cite{Dud20a}. The trend component is shaped by country-specific factors, including economic growth rates, industrial activity, and climate variations~\cite{Hor05}. Seasonal fluctuations largely reflect regional climatic influences and weather patterns~\cite{Apa12}, along with the heterogeneous makeup of electricity consumers within each nation. These seasonal patterns, with their regular cycles, offer critical insights into recurring demand behaviors. 

Despite these identifiable patterns, accurate forecasting is challenged by several disruptive factors. Unanticipated economic events, extreme weather episodes, and changes in political or regulatory policies introduce considerable volatility and uncertainty into demand patterns~\cite{Dog16}. Such disruptions amplify the complexity of forecasting future electricity demand and complicate the extraction of reliable signals from historical data.

Let ${Y} = \{y_1, \dots, y_T\}$ represent a monthly electricity demand time series, where $y_t \in \mathbb{R}$ denotes the observed value at time step $t$ and $T$ is the total length of the time series. The goal of MTLF is to predict the future values of the time series over a forecasting horizon $H$, i.e. values ${\mathbf{y}}_T = [ {y}_{T+1}, \dots, {y}_{T+H} ]$. 
To achieve this, the model takes as input a lookback window of length $w \le T$, which contains the most recent observations up to $y_T$. This lookback window, denoted by $\mathbf{x}_T = [y_{T-w+1}, \dots, y_T]$, serves as the historical context for the forecast. In this study, no exogenous variables are incorporated as additional inputs, so the approach focuses on univariate MTLF, relying solely on the demand series itself.

The forecasting model is expressed as $f(\mathbf{x}; \Theta)$, where $\Theta$ represents the model’s parameters and hyperparameters. It is trained on historical data, $\{(\mathbf{x}_t, \mathbf{y}_t)\}_{t \in \Xi}$, where $\Xi$ is a set of input-output pairs selected from past observations. Model training aims to minimize the difference between the actual and predicted values, typically using a predefined loss function. Our model is trained in a global (cross-learning) setting, utilizing monthly electricity demand time series data from multiple countries. This approach requires equipping the model with mechanisms to handle differences in scale and variability inherent to each time series, ensuring adaptability to diverse demand patterns across countries.

Performance evaluation of the model typically relies on metrics such as Mean Absolute Percentage Error (MAPE), Root Mean Square Error (RMSE), or other domain-relevant measures, which assess both accuracy and robustness in predicting electricity demand.

\section{Forecasting Model} \label{FM}

N-BEATS (Neural Basis Expansion Analysis for time-series Forecasting) is a deep neural network model designed for time-series forecasting \cite{Ore19}. Unlike many other forecasting models, N-BEATS frames forecasting as a non-linear multivariate regression task rather than a sequence-to-sequence problem. Its design philosophy centers around conceptual simplicity, flexibility, and high performance without requiring domain-specific feature engineering or time-series decomposition. N-BEATS offers two configurations: the \emph{generic} version, which optimizes performance without requiring interpretable outputs, and the \emph{interpretable} version, which decomposes the forecast into human-understandable components such as trend and seasonality. This decomposability makes N-BEATS suitable for applications where interpretability is crucial. In this study, we use the generic version. N-BEATS was adapted for MTLF in~\cite{Ore21}. Here, we present further development of the model, which we henceforth refer to as N-BEATS*, including the refinement of the time-series processing within network blocks and a new loss function. These modifications enhance the model by addressing the unique challenges associated with the MTLF problem.

\begin{figure}[t]
\centering
\includegraphics[width=\textwidth]{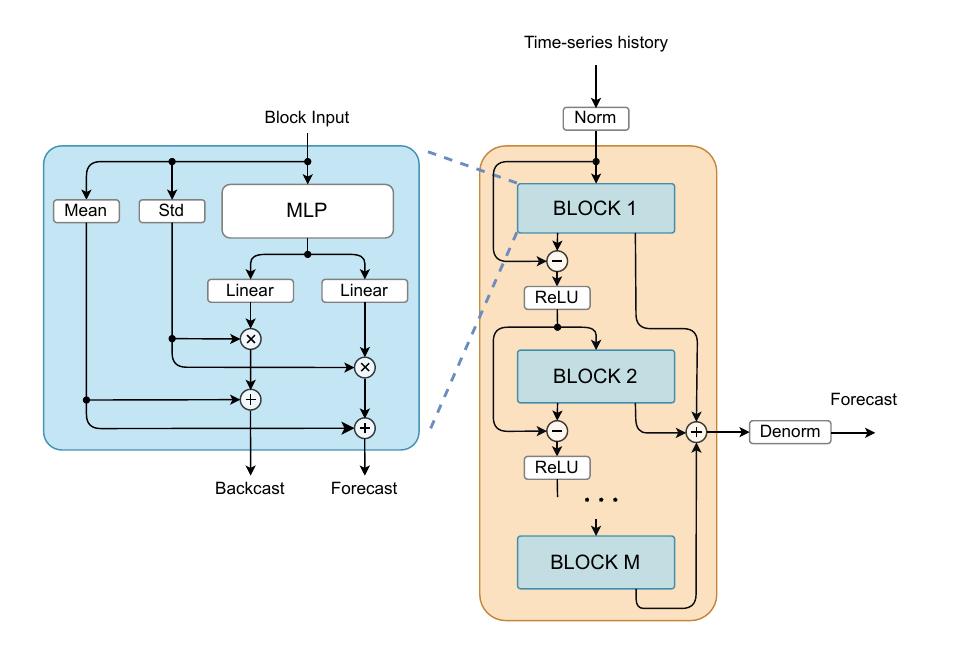}
\caption{Architecture of the N-BEATS* model.} \label{fig:nbeats}
\end{figure}

\subsection{N-BEATS* Architecture}

The N-BEATS* model is illustrated in Fig.~\ref{fig:nbeats}. It is composed of the stack of blocks connected via residual connections, forming a deep hierarchy. This allows the model to capture complex, nonlinear relationships in time-series data, leveraging both forward and backward residual links. Residual connections facilitate better gradient flow during training, allowing effective stacking of numerous blocks. The stacked block architecture enables the model to iteratively refine its predictions, making it highly effective at capturing patterns such as trends and seasonality. The forecasts generated by each block are summed to produce the final forecast. 
Each block in the N-BEATS* model consists of fully connected layers followed by non-linear activations, typically ReLU (\textsc{MLP} component in Fig.~\ref{fig:nbeats}). 
The architecture features a dual-path mechanism (a fork) that predicts both a forecast for future values (termed as the \emph{forecast} path) and a reconstruction of past values (termed as the \emph{backcast} path). 

We enhance the standard N-BEATS block by introducing a destandardization step for the forecast and backcast vectors (represented as \textsc{Mean} and \textsc{Std} components in Fig.~\ref{fig:nbeats}). This transformation simplifies the task of the backcast/forecast fork by allowing it to predict sequences that are standardized, with zero mean and unit variance, rather than sequences varying in level and variance. As a result, the forecast and backcast vectors predicted for different countries differ only in shape, not in level or variance. The appropriate level and variance are subsequently restored through the \textsc{Mean} and \textsc{Std} components, making the model more robust and consistent across diverse time-series. This adjustment is particularly important given that the model is trained in cross-learning mode, meaning it learns simultaneously from time-series data across multiple countries.

Note that the residual connections between blocks are transformed using the ReLU function, eliminating negative values from the inputs to the next block. Additionally, the model’s input 
x-vectors are normalized (shown as the \textsc{Norm} component in Fig.~\ref{fig:nbeats}) by dividing each by its maximum value. This normalization unifies the input vectors, ensuring that each has a maximum value of 1. Consequently, the model operates on normalized data and generates consistent forecasts, which are then denormalized (via the \textsc{Denorm} component in Fig.~\ref{fig:nbeats}) to restore the appropriate scale (the maximum value of the original input vector).

The N-BEATS* model can be expressed by the following equations:
\begin{align}  \label{eq1}
\textsc{Input normalization:} \quad\quad \mathbf{x}^{(1)} &= \frac{\mathbf{x}}{\max{(\mathbf{x}})} \nonumber\\
\textsc{For Block}\ m = 1\ldots\text{\emph{M}:} \quad \mathbf{h}^{(m)} &= \textsc{FC}(\mathbf{x}^{(m)}) \nonumber\\ 
\hat{\mathbf{x}}^{(m)} &=\textsc{Linear}(\mathbf{h}^{(m)}) \cdot \textsc{Std}(\mathbf{x}^{(m)}) + \textsc{Mean}(\mathbf{x}^{(m)}) \nonumber\\
\hat{\mathbf{y}}^{(m)} &= \textsc{Linear}(\mathbf{h}^{(m)}) \cdot \textsc{Std}(\mathbf{x}^{(m)}) + \textsc{Mean}(\mathbf{x}^{(m)}) \nonumber\\
\mathbf{x}^{(m+1)}&=\textsc{ReLU}(\mathbf{x}^{(m)} - \hat{\mathbf{x}}^{(m)}) \nonumber\\
\textsc{Output:} \quad\quad\quad \hat{\mathbf{y}} &= \max{(\mathbf{x}}) \cdot \sum_{m=1}^{M}{\hat{\mathbf{y}}^{(m)}} 
\end{align}

\subsection{Loss Function}

Given that MAPE is a well-established performance metric for electricity load forecasting,~\cite{Ore21} introduces the pinball-MAPE loss function, which aligns training and evaluation metrics and provides a leverage for controlling forecast bias:
\begin{align}
\textsc{pMAPE}(\mathbf{y},\hat{\mathbf{y}},\tau) = \frac{1}{N\cdot H} \sum_{i=1}^N\sum_{j=1}^H
    \begin{cases}
         \tau \frac{(y_{i,j} - \hat y_{i,j})}{y_{i,j}} & \text{if} \quad y_{i,j} \geq \hat y_{i,j}, \\
         (1-\tau) \frac{(\hat y_{i,j} - y_{i,j})}{y_{i,j}}        & \text{otherwise}.
    \end{cases} 
 \nonumber
\end{align}
Here $N$ represents the number of forecasted sequences evaluated by the loss function (e.g. the batch size), $H$ is the forecast horizon, $\tau$ defines the quantile probability ($\tau=0.5$ corresponds to the median). In this study, we extend the pinball-MAPE (\textsc{pMAPE}) by adding the \textsc{nMSE} (normalized MSE) term, formulated as follows:
\begin{align}
\textsc{nMSE}(\mathbf{y},\hat{\mathbf{y}})= \frac{1}{N\cdot H} \sum_{i=1}^N\sum_{j=1}^H \frac{(y_{i,j} - \hat y_{i,j})^2}{\textsc{Var}(\mathbf{y}_i)},
\end{align}
where \textsc{Var} denotes the variance. This can be interpreted as the ratio of two MSEs: one for the model-generated forecast and one for a baseline forecast given by the mean of the target sequence. Thus, the normalized MSE equals 1 when the model’s squared error matches the error of the mean-based baseline.

\begin{figure}[t]
\centering
\includegraphics[width=0.40\textwidth]{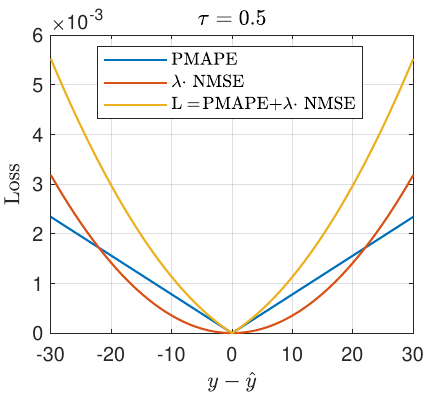}
\includegraphics[width=0.40\textwidth]{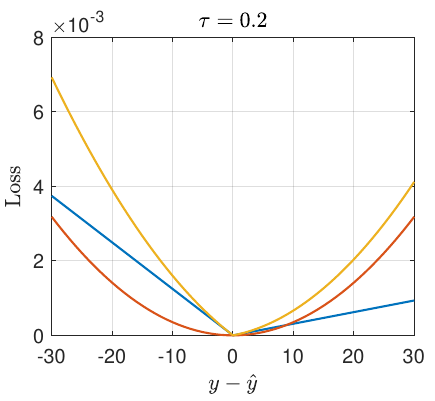}
\caption{Example loss functions for $\tau=0.5$ and $\tau=0.2$.} \label{fig:loss}
\end{figure}

The proposed loss combines the two losses described above with coefficient $\lambda \geq 0$ controlling the relative influence of \textsc{nMSE} and \textsc{pMAPE} over the overall loss function:
\begin{gather}  \label{loss}
L(\mathbf{y},\hat{\mathbf{y}},\tau)= \textsc{pMAPE}(\mathbf{y},\hat{\mathbf{y}},\tau) + \lambda \cdot \textsc{nMSE}(\mathbf{y},\hat{\mathbf{y}}).
\end{gather}
Figure~\ref{fig:loss} shows the components of the proposed loss function. For $\tau=0.5$, \textsc{pMAPE} is symmetric, meaning that positive and negative deviations are treated equally. When 
$\tau \neq 0.5$, however, the pinball function becomes asymmetric, which can aid in correcting forecast bias. If the model produces biased forecasts, slightly adjusting $\tau$ can help reduce this bias. Thus, the asymmetric loss function provides flexibility for bias mitigation. The \textsc{nMSE} component of the loss function penalizes larger errors more heavily than the \textsc{pMAPE} component, emphasizing the reduction of significant deviations. The combined effect of these two components ensures a balanced approach, improving both accuracy and robustness in forecasting.

\section{Experiments} \label{EX}

In this section, we evaluate the proposed N-BEATS* model on the MTLF task, benchmarking its performance against a range of models, including classical statistical methods, machine learning techniques, and hybrid approaches.

\subsection{Data}

This study uses real-world data from the ENTSO-E platform (\url{www.entsoe.eu}), comprising monthly electricity consumption time series for 35 European countries. The longest series span from 1991 to 2014 (11 countries), while others cover shorter periods: 17 years (6 countries), 12 years (4 countries), 8 years (2 countries), and 5 years (12 countries). Visualizations are available in \cite{Dud20a}, \cite{Ore21}. Each country’s time series exhibits unique dynamics and characteristics, including trends, seasonality, and random fluctuations. This diversity offers a robust testing ground for evaluating the ability of forecasting models to capture and predict complex electricity demand patterns.

\subsection{Optimization, Training and Evaluation Setup}

The dataset was divided into three subsets for model development and evaluation. The test set consists of the final twelve months (2014) of each time series, while the validation set contains the preceding twelve months (2013). The training set encompasses all remaining historical data prior to 2013.
We employed a two-stage process for model development. First, we used the training and validation subsets to optimize hyperparameters. Then, after determining the optimal hyperparameters, we merged the training and validation sets to train the final model. Finally, we evaluate the model's performance on the held-out test set. This split ensures proper temporal separation between model development and final evaluation, while maximizing the data available for the final model training.

The optimization and training methodology followed the approach established for the N-BEATS* predecessor in \cite{Ore21}. Hyperparameter configurations are detailed in Table \ref{tabH}.
To address the varying lengths of time series in our dataset, we implemented weighted stratified sampling during training, as described in \cite{Ore21}. This sampling strategy ensures balanced representation by equalizing the frequency at which training samples from both shorter and longer series contribute to the loss function adjustment. Given the stochastic nature of N-BEATS*, we conducted extensive experimentation to ensure robust results. For each reported metric, we averaged the outcomes across 100 independent trials. Each trial consisted of an ensemble of 64 models, randomly selected from a pool of 1024 trained models using bootstrap sampling. Model diversity within the ensemble was achieved through random initialization of parameters and varying the sequence of training batches.

\begin{table}[t]
    \centering
    \caption{N-BEATS* hyperparameter settings.}
    \label{tabH}

    \begin{tabular}{lcc}
        \toprule
        Hyperparameter & Value  & Search grid \\ 
        \midrule
        Pinball $\tau$ & 0.35 & [0.3, 0.35, ..., 0.6] \\
        NMSE weight $\lambda$ & 0.35 & [0.3, 0.35, ..., 0.6] \\
        FC width & 512 & [256, 512, 1024]\\ 
        \#blocks ($M$) & 6 & [1, 2, 3, 6, 12]	  \\
        \#FC layers  & 3 & [2, 3, 4]  \\
        Sharing & True & [True, False]	  \\
        Lookback period ($w$, months) & 12 & [6, 9, 12, 24]  \\
        Epochs	& 20 & 20	\\ 
        Batches per epoch & 100 & [50, 100, 150] \\
        Batch size & 256 & [128, 256, 512, 1024]	\\
        Optimizer & Adam & Adam \\
        Learning rate & 0.001 & 0.001 \\
        Ensemble size & 64 & 64 \\
        \bottomrule
    \end{tabular}
\end{table}


\subsection{Baseline Models}

In our comparative studies, we evaluate the performance of N-BEATS* against several baseline models. The optimization and training procedures for these baseline models are similar to those used for N-BEATS*, with hyperparameter settings described in detail in \cite{Ore21}.

\begin{itemize}

    \item ARIMA and ETS: Classical statistical models, implemented using the \texttt{auto.arima} and \texttt{ets} functions from the R package \texttt{forecast} ~\cite{Hyn20}. Both models leverage the Akaike information criterion (AICc) to automatically determine the optimal model structure and order.
    \item $k$-NNw+ETS, FNM+ETS, N-WE+ETS, GRNN+ETS: Hybrid models that combine either $k$-nearest neighbor weighted regression, fuzzy neighborhood model, Nadaraya–Watson estimator, or general regression neural network for seasonal component forecasting, with ETS for trend and dispersion forecasting~\cite{Dud20a}.
    \item MLP: A perceptron with a single hidden layer and sigmoid non-linearities~\cite{Pel19b}.
    \item ANFIS: A standard adaptive neuro-fuzzy inference system~\cite{Pel18}.
    \item LSTM: A standard LSTM model~\cite{Pel20}.
    \item ETS+RD-LSTM: A hybrid model that combines ETS, an advanced LSTM architecture with residual and dilated connections~\cite{Dud21}.
    \item N-BEATS: The predecesor of N-BEATS* described in \cite{Ore21}.

\end{itemize}

\subsection{Data Processing by N-BEATS*}

\begin{figure}[t]
\centering
\includegraphics[width=0.45\textwidth]{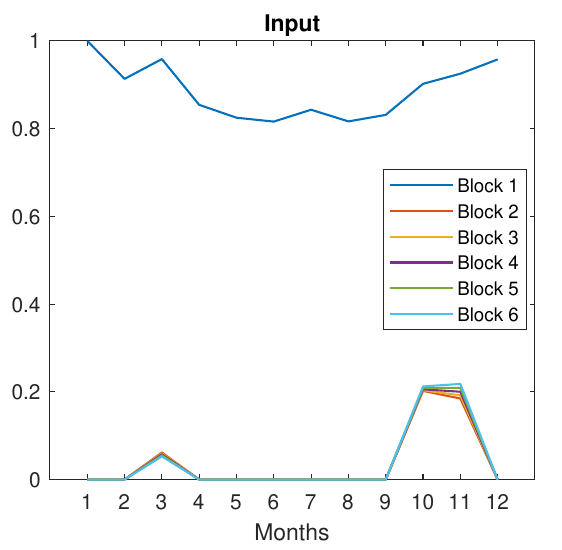}
\includegraphics[width=0.45\textwidth]{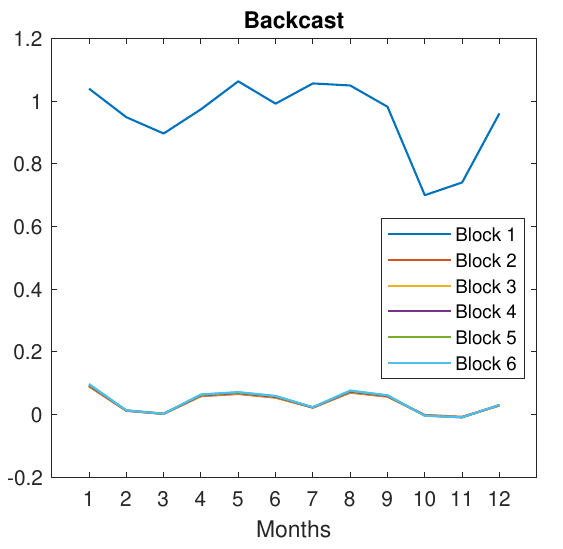}
\includegraphics[width=0.45\textwidth]{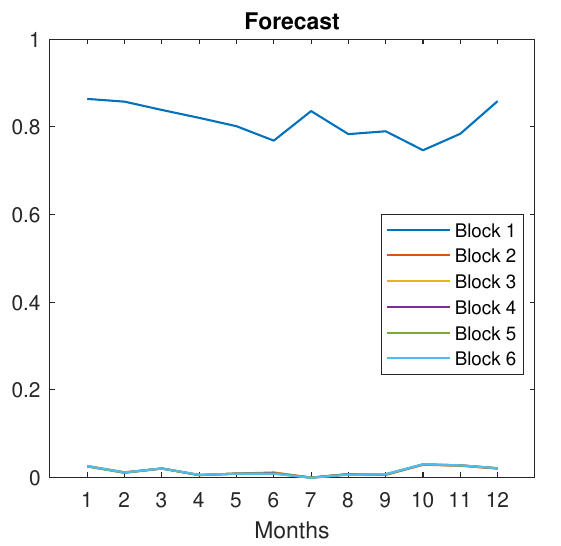}
\includegraphics[width=0.45\textwidth]{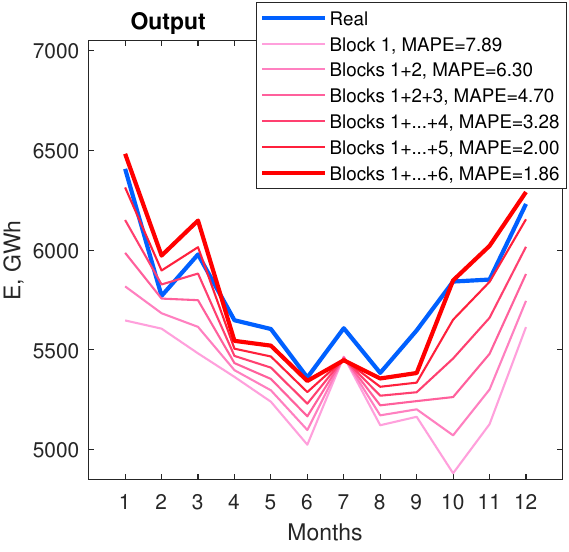}
\caption{Data processing by N-BEATS -- example showing test data for Austria.} \label{fig2}
\end{figure}

Fig. \ref{fig2} illustrates the data processing flow across successive blocks. The input to \textsc{Block} 1 is a normalized yearly demand curve for a given country. This block produces a forecast vector, which forms the primary component of the final forecast, and a backcast vector. After subtracting the backcast vector from the input to \textsc{Block} 1 and applying the ReLU function, the resulting vector becomes the input to the second block. As shown in Fig. \ref{fig2}, the input vectors to \textsc{Blocks} 2-6 are highly similar, as are the forecast and backcast vectors produced by these blocks. The forecast vectors generated by \textsc{Blocks} 2-6 serve as incremental adjustments to the main forecast produced by \textsc{Block} 1. With each successive block, the refined forecast curve increasingly aligns with the target curve (see the bottom-right plot in Fig. \ref{fig2}). As more blocks are added, MAPE decreases progressively, from 7.89 to 1.86 in the example shown.

\subsection{Results} \label{RE}

Table \ref{tab1} presents the forecasting metrics averaged across 35 countries: the median absolute percentage error (MedAPE), mean absolute percentage error (MAPE), interquartile range of APE (IQR APE), root mean square error (RMSE), and mean percentage error (MPE). As shown in this table, our proposed N-BEATS* model achieves the lowest errors across several metrics: it yields the lowest Median APE, MAPE, and RMSE, and also produces forecasts with the least dispersion, as indicated by the lowest IQR APE. Compared to N-BEATS, N-BEATS* reduces MAPE by up to 9\% and RMSE by 1.6\%. To statistically confirm the improved accuracy of N-BEATS* over N-BEATS, we applied the Diebold-Mariano test, which assesses the equality of forecasting accuracy between two models under general assumptions. The test statistic, which follows an asymptotic standard normal distribution, was calculated as $-3.05$. This value is below the critical z-value of $-2.576$ for a significance level of $\alpha=
0.01$, indicating statistically significant superiority of N-BEATS*. Furthermore, in \cite{Ore21}, we demonstrated that N-BEATS outperforms each baseline model listed in Table \ref{tab1} at the $\alpha=0.01$ significance level. Together, these results confirm the overall superiority of N-BEATS* over both N-BEATS and all baseline models.

\begin{table}[]
\caption{Forecasting metrics.}
	\label{tab1}
	\setlength{\tabcolsep}{6.4pt}
	\centering
	\begin{tabular}{lcccccc}
		\toprule
		Model       & \multicolumn{1}{c}{$\textsc{$\medape$}$} & \multicolumn{1}{c}{$\mape$} & \multicolumn{1}{c}{$\iqr$ $\ape$} & \multicolumn{1}{c}{$\rmse$} & \multicolumn{1}{c}{$\mpe$}  \\ \midrule \vspace{0.1cm}
		
		ARIMA	&	3.32	&	5.65	&	5.24 & 463 & --2.35	 \\ 
		\vspace{0.1cm}
		ETS	&	3.50	&	5.05	&	4.80	& 374 & --1.04 \\
		\vspace{0.1cm}
		k-NNw+ETS	&	2.71	&	4.47	&	3.52	&	327 & --1.25	  \\ \vspace{0.1cm}
		FNM+ETS	&	2.64	&	4.40	&	3.46	&	321 & --1.26	\\ \vspace{0.1cm}
		N-WE+ETS	&	2.68	&	4.37	&	3.36	&	320 & --1.26	\\ \vspace{0.1cm}
		GRNN+ETS	&	2.64	&	4.38	&	3.51	&	324 & --1.26	\\ \vspace{0.1cm}
		MLP	&	2.97	&	5.27	&	3.84	&	378  & --1.37 \\ \vspace{0.1cm}
		ANFIS	&	3.56	&	6.18	&	4.87	&	488 & --2.51	 \\ \vspace{0.1cm}
		LSTM	&	3.73	&	6.11	&	4.50	&	431 & --3.12	 \\ \vspace{0.1cm}
		ETS+RD-LSTM	&	2.74	&	4.48	&	3.55	&	347 & --1.11	 \\  \vspace{0.1cm}	 
		N-BEATS	&	{2.55}	&	{3.78}	&	{3.30}	&	{309} & \textbf{--0.34}	 \\
            N-BEATS*	&	\textbf{2.20}	&	\textbf{3.44}	&	\textbf{3.29}	&	\textbf{304} & {0.56}	 \\
		\bottomrule
	\end{tabular}
\end{table}

\begin{figure}[t]
\centering
\includegraphics[width=1\textwidth]{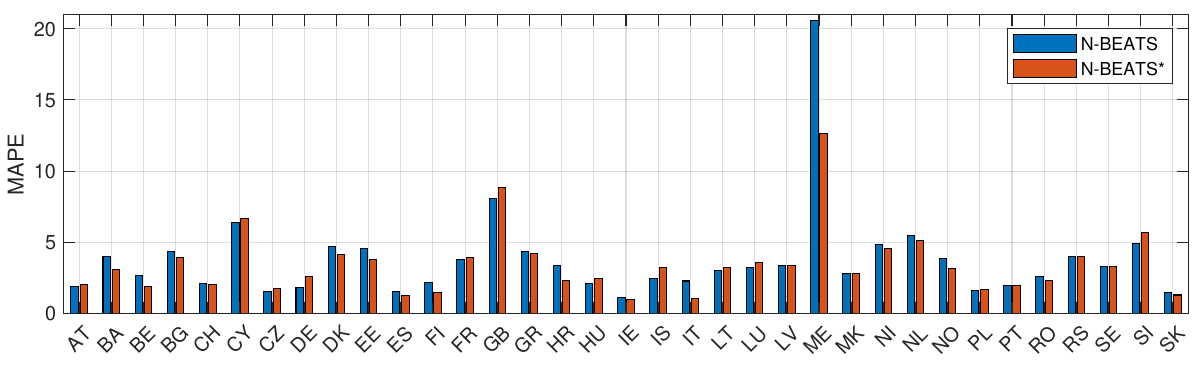}
\caption{MAPE for individual countries.} \label{fig3}
\end{figure}

\begin{figure}[t]
\centering
\includegraphics[width=0.32\textwidth]{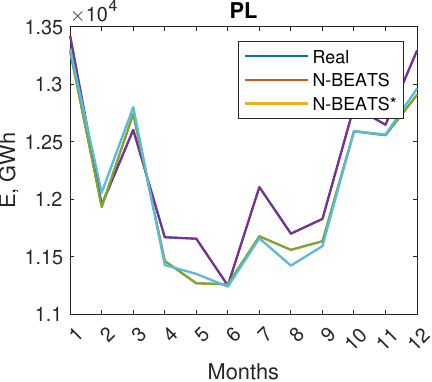}
\includegraphics[width=0.32\textwidth]{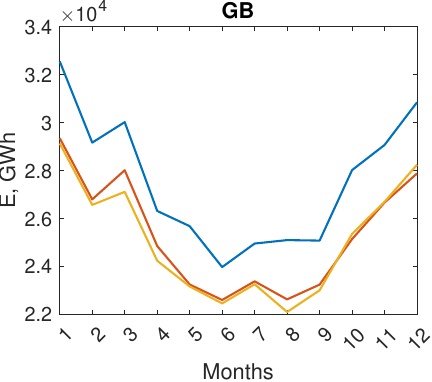}
\includegraphics[width=0.32\textwidth]{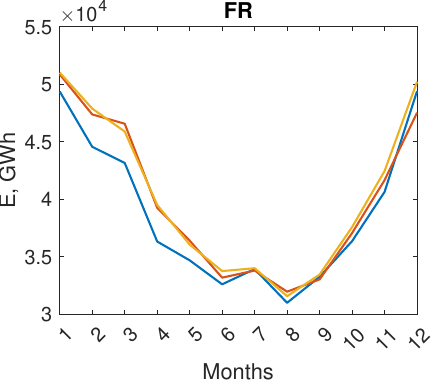}
\includegraphics[width=0.32\textwidth]{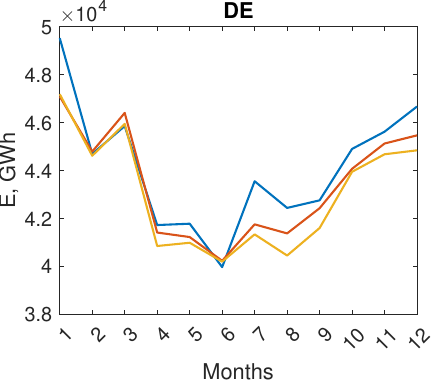}
\includegraphics[width=0.32\textwidth]{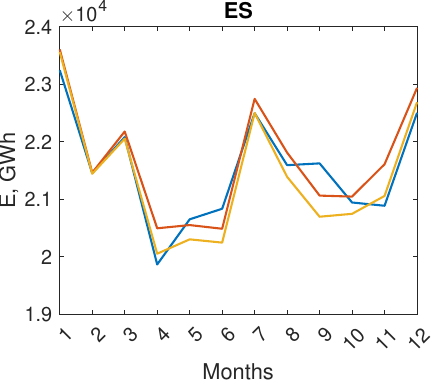}
\includegraphics[width=0.32\textwidth]{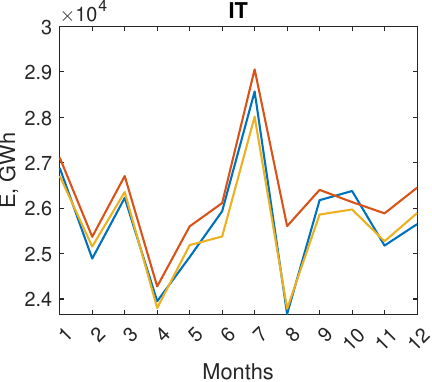}
\includegraphics[width=0.32\textwidth]{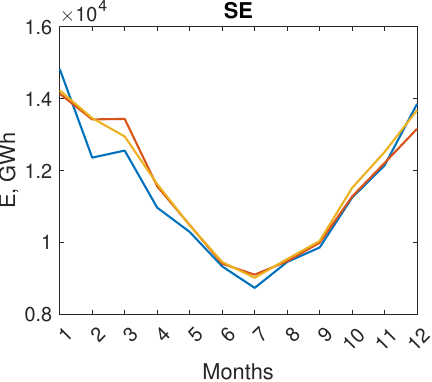}
\includegraphics[width=0.32\textwidth]{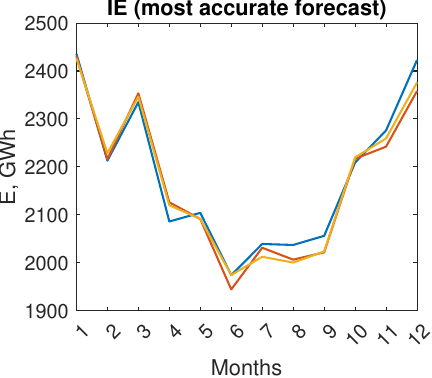}
\includegraphics[width=0.32\textwidth]{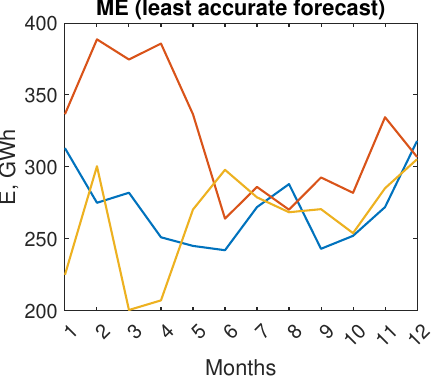}
\caption{Forecasts for selected countries.} \label{fig5}
\end{figure}

Fig. \ref{fig3} shows a country-by-country comparison of MAPE between N-BEATS and N-BEATS*. In this analysis, N-BEATS* outperforms N-BEATS in 20 out of 35 cases, while N-BEATS has a slight edge in 15 cases. The largest improvement, observed for Montenegro (ME), reaches 38.5\%, as further illustrated in Figure \ref{fig5}. 

Forecast examples generated by N-BEATS* and N-BEATS are shown in Fig. \ref{fig5}. Note that the forecasts for Great Britain (GB) are underestimated, a result of an unexpected rise in demand during the forecast year despite a preceding downward trend. Conversely, for France (FR), an opposite trend led to a slight overestimation. The most accurate forecast, with a MAPE of 0.97\%, is for Ireland (IE), while the least accurate, with a MAPE of 12.67\%, is for Montenegro (ME), influenced by an outlier in the series in the year preceding the forecast (see Fig. \ref{fig0}).

\begin{figure}[t]
\centering
\includegraphics[width=0.60\textwidth]{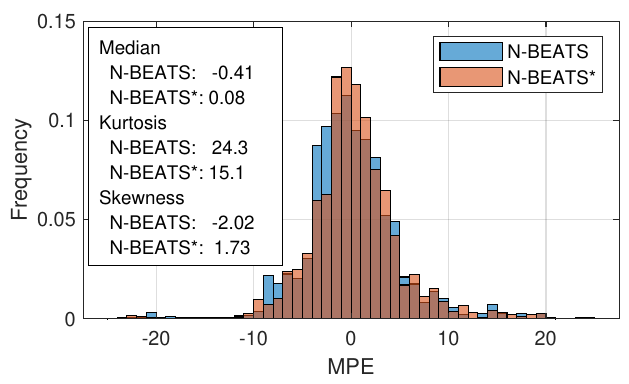}
\caption{Histograms of MPE.} \label{fig6}
\end{figure}

MPE, shown in Table~\ref{tab1}, provides insight into the forecast bias of both the proposed and baseline models. Notably, N-BEATS* is the only model that produces positively biased forecasts, indicating a tendency for underprediction, whereas all other models tend to overpredict. N-BEATS* achieves one of the lowest biases, with MPE of 0.56\%, closely following N-BEATS, which has MPE of $-0.34\%$. Fig. \ref{fig6} illustrates the MPE distributions for these two models, where N-BEATS* shows a higher concentration of forecasts with MPE values close to zero, indicating more frequent low-bias predictions.

In terms of skewness, N-BEATS* exhibits a skewness of 1.73, suggesting a more balanced error profile with a mild tendency toward overprediction. In contrast, N-BEATS has a negative skewness, indicating a propensity for occasional, larger overpredictions.

Both models show high kurtosis values, suggesting that their error distributions include more extreme deviations or outliers. However, the reduction in both kurtosis and skewness from N-BEATS to N-BEATS* is beneficial, as it implies that N-BEATS* produces a more consistent error pattern with fewer significant directional biases. This improvement contributes to the overall reliability of the forecasts.

\subsection{Ablation Study}

In this section, we perform an ablation study to assess the impact of various modifications to the N-BEATS* model on its forecasting performance. We evaluate the following simplified variants of N-BEATS*:

\begin{itemize}
    \item [\textbf{noL2}] N-BEATS* without the \textsc{nMSE} term in loss function \eqref{loss}.
    \item [\textbf{noVar}] N-BEATS* without normalizing the $L_2$ component by the target series variance (omitting $\textsc{Var}(\mathbf{y}_i)$ in \eqref{loss}).
    \item [\textbf{noDestd}] N-BEATS* without the destandardization of the forecast and backcast outputs (removal of the \textsc{Mean} and \textsc{Std} components, as depicted in Fig. 1).
    \item [\textbf{noReLU}] N-BEATS* without the ReLU activation function applied to the inputs of blocks 2 to $M$ (omission of the \textsc{ReLU} components in Fig. 1). 
\end{itemize}

Table \ref{tabAb} compares forecasting errors for the full model and its reduced variants. Based on the results from this table, the most significant performance degradation occurs when the \textsc{nMSE} term and its normalization by variance are removed, with the noVar variant showing the highest error rates. In contrast, the removal of destandardization and the ReLU activation function has a more moderate impact, with similar performance declines across these variants. Proper formulation of the loss function, including the \textsc{nMSE} term with variance normalization, is crucial for maintaining forecasting accuracy in the N-BEATS* model.

\begin{table}[]
	\setlength{\tabcolsep}{5pt}
	\caption{Errors for the full and reduced N-BEATS* model.}
	\begin{center}
	\begin{tabular}{ccccccc}
		\toprule
		& N-BEATS*  & noL2   & noVar & noDestd  & noReLU\\
		\midrule    
    MAPE  & \textbf{3.44} & 4.25 & 4.50 & 4.01 & 4.01 \\
    RMSE  & \textbf{304} & 322 & 332 & 314 & 309 \\
 		\bottomrule
	\end{tabular}
	\label{tabAb}
	\end{center}
\end{table}

\section{Discussion} \label{DS}

The results presented in Section \ref{RE} demonstrate that N-BEATS* exhibits superior performance compared to an array of forecasting methods, including statistical models (ARIMA and ETS), classical machine learning techniques (MLP, ANFIS, and LSTM), and hybrid approaches. Notably, N-BEATS* consistently achieves the lowest Median APE, MAPE, and RMSE values, while simultaneously demonstrating the lowest dispersion of forecast errors, as reflected by its IQR APE.

\subsection{Sophisticated Architecture for Raw Time Series}

Several factors contribute to the success of N-BEATS*. The model's inherent ability to effectively handle raw time series data without requiring decomposition or preprocessing is a significant advantage. Many statistical and machine learning methods struggle with non-stationarity, non-linear relationships, and seasonal variations, often necessitating preliminary steps such as differencing, detrending, deseasonalization, or decomposition. In contrast, N-BEATS* leverages its sophisticated architecture, incorporating backward and forward residual links, a deep stack of fully connected layers, forecast and backcast paths, and hierarchical aggregation of partial forecasts, to adeptly process raw time series data. This architectural advantage, in conjunction with the final ensembling process, results in highly accurate forecasts.

\subsection{Ensembling}

The ensemble approach is a key factor in the model's success, significantly enhancing its forecasting accuracy and reliability. By aggregating predictions from multiple ensemble members (64 in our implementation), the model reduces the variability and overfitting often associated with single-model predictions. In N-BEATS*, variability among individual ensemble members arises from random initialization and random batch selection during training. This diversity strengthens the ensemble’s robustness and generalization, resulting in consistently more accurate and dependable forecasting outcomes.

\subsection{Cross-learning}

The cross-learning approach employed during the training of N-BEATS* is another critical factor in its success. By training on multiple time series simultaneously, the model captures shared features and components, accelerating learning and optimization, particularly crucial for complex deep learning models with numerous parameters and hyperparameters. While other models, except for ETS+RD-LSTM, undergo separate training and optimization for each time series, N-BEATS* leverages the collective information embedded within multiple time series to achieve superior performance.

\subsection{Key Innovations}

The introduction of the destandardization component in the block architecture simplifies forecasting tasks by harmonizing time series with varying scales. This innovation enables the model to handle cross-learning more effectively, leveraging shared patterns across multiple time series. As a result, N-BEATS* demonstrates improved accuracy and robustness, particularly in datasets characterized by non-stationarity and high variability.

The novel loss function, combining pinball loss based on MAPE with normalized MSE, proves instrumental in achieving good results. This hybrid formulation balances the benefits of both 
\textsc{pMAPE} and \textsc{nMSE} loss terms, controlling forecast bias while maintaining high accuracy. The ability of the pinball loss component to forecast quantiles further enhances the model's flexibility, allowing it to be tailored for specific operational requirements in power system planning.

\subsection{Practical Implications}

The enhanced performance of N-BEATS* translates directly to practical benefits for power system operators. Improved accuracy reduces the risk of under- or over-allocation of resources, while the model's bias control ensures reliability in planning scenarios. The ability to train the model on raw time series without requiring domain-specific preprocessing makes N-BEATS* particularly suited for large-scale deployment across diverse regions.

\subsection{Limitations}

While N-BEATS* demonstrates clear advantages, certain limitations warrant further exploration and refinement:

\begin{itemize}
    \item \textbf{Generalizability to Other Forecasting Problems}:
    This study focuses on applying N-BEATS* to MTLF, leveraging its strengths in mid-term time horizons. However, further investigation is needed to evaluate its performance on problems with different characteristics, such as varying data frequencies, shorter time horizons, or specific domain requirements. For instance, the model's effectiveness in short-term forecasting tasks with triple seasonality and hourly resolutions has not yet been explored. While the original N-BEATS was designed for general time series forecasting, adapting N-BEATS* to broader contexts remains an open question.
    \item \textbf{Limited Use of Exogenous Variables}:
    The current implementation of N-BEATS* is designed for univariate time series, relying solely on historical demand data. Many forecasting tasks, such as weather prediction or sales forecasting, require the integration of exogenous variables to improve accuracy. The absence of support for external inputs limits the applicability of N-BEATS* to problems where external factors play a significant role in influencing outcomes.
    \item \textbf{Computational Cost}:
     While N-BEATS* maintains a relatively efficient design for deep learning models, its deep architecture with stacked blocks and ensembling approach still demands substantial computational resources for training and optimization. This complexity could pose challenges when scaling to very large datasets or deploying in applications with strict real-time constraints. Strategies to simplify the architecture without compromising performance may help address this limitation.
    \item \textbf{Interpretability}:
     Although an interpretable variant of N-BEATS exists, this study employs the generic version, prioritizing performance over transparency. The impact of the introduced modifications, such as the new loss function and the modified block architecture, on the interpretability of the network has not been assessed. A lack of transparency could hinder adoption in domains where decision-making requires clear explanations for predictions.
     \item \textbf{Sensitivity to Data Quality}:
    As a purely data-driven model, N-BEATS* heavily relies on the quality and representativeness of the training data. Issues such as outliers, missing values, or inconsistencies can negatively affect its accuracy and robustness.
    \item \textbf{Sensitivity to Hyperparameter Tuning}:
    N-BEATS* performance depends on carefully tuned hyperparameters, such as the pinball quantile probability ($ \tau $), the loss function’s NMSE weight ($ \lambda $), and the lookback window size ($ w $). Determining optimal values for these parameters can require extensive experimentation, which might limit the ease of use and adaptability of the model in new contexts.
     
    \item \textbf{Potential for Overfitting}:
     The deep architecture of N-BEATS*, with numerous trainable parameters, introduces a risk of overfitting, particularly when trained on limited datasets. While cross-learning and ensembling techniques help mitigate this risk, careful application of regularization methods and robust validation procedures remain crucial to ensure the model generalizes well to unseen data. 
    \item \textbf{Dependency on Cross-Learning}:
    N-BEATS* benefits from cross-learning by identifying shared patterns across multiple time series. While this approach is highly effective for MTLF, it may perform less effectively on datasets consisting of isolated or highly heterogeneous time series. For such cases, alternative training paradigms or domain-specific adaptations may be necessary to achieve optimal performance.
    \item \textbf{Dependence on Ensembling:}
    The high accuracy of N-BEATS* largely stems from leveraging an ensemble of multiple models. Individual models typically produce less accurate forecasts, making ensemble averaging a critical component for improved performance. However, this dependency significantly increases computational demands, potentially posing challenges in resource-constrained environments.
    
\end{itemize}

Addressing these limitations through further research and innovation could broaden the applicability and impact of N-BEATS*, making it a more versatile and robust tool for time series forecasting.

\subsection{Future Research Directions}

Future work could focus on extending N-BEATS* to incorporate exogenous variables, such as economic indicators or weather data, to enhance its MTLF predictive accuracy. Additionally, leveraging the interpretable version of N-BEATS for MTLF could provide valuable insights for decision-makers by decomposing forecasts into trend and seasonal components, improving transparency and actionable understanding. Another promising avenue is extending N-BEATS* for probabilistic forecasting. In \cite{Smy24}, we proposed an any-quantile variant of N-BEATS -- a novel approach for distributional forecasting capable of predicting arbitrary quantiles -- which could be adapted and applied to MTLF scenarios.

Other promising research directions include: 
\begin{itemize}
    \item Adapting the model for diverse time series structures: Extending N-BEATS* to handle irregular or high-frequency data, such as those found in sensor networks or financial markets.
    \item Exploring applications beyond MTLF: Applying N-BEATS* to domains like financial forecasting or healthcare, incorporating domain-specific modifications to address unique challenges in these fields.
    \item Improving computational efficiency and interpretability: Investigating strategies to reduce computational costs for scalability and enhance interpretability to meet the demands of applications requiring transparent and explainable predictions.
\end{itemize}

\section{Conclusion} \label{CN}

This paper presents N-BEATS*, an enhanced version of the N-BEATS model tailored for mid-term load forecasting. By introducing a novel block architecture with a destandardization component and a hybrid loss function combining pinball-MAPE loss and normalized MSE, N-BEATS* achieves significant improvements in accuracy. Empirical evaluation on real-world data from 35 European countries demonstrates that N-BEATS* outperforms its predecessor and state-of-the-art forecasting methods, providing robust and reliable predictions. The proposed enhancements improve forecasting performance and address key MTLF challenges, such as managing diverse time series. While N-BEATS* demonstrates clear advantages, future work could extend its applicability to other domains, incorporate exogenous variables, and enhance computational efficiency. Overall, N-BEATS* marks a significant advancement in time series forecasting, particularly for the complex demands of MTLF.






\bibliographystyle{elsarticle-harv}
\bibliography{Arxiv_version}

\end{document}